\documentclass[journal]{IEEEtran}

\ifCLASSOPTIONcompsoc
  \usepackage[nocompress]{cite}
\else
  \usepackage{cite}
\fi

\usepackage{amsmath}
\DeclareMathOperator*{\argmax}{argmax}

\usepackage{algorithm}
\usepackage{algorithmic}
\usepackage{graphicx}
\usepackage{subcaption}
\usepackage{comment}
\usepackage{url}
\usepackage{amssymb}
\usepackage{booktabs}
\usepackage{multirow}
\usepackage[dvipsnames]{xcolor}

\begin{document}

\title{Deep Echo State Q-Network (DEQN) and Its Application in Dynamic Spectrum Sharing for 5G and Beyond}

\author{Hao-Hsuan~Chang, Lingjia~Liu, and Yang Yi
\thanks{The authors are with the Bradley Department of Electrical and Computer Engineering, Virginia Tech, Blacksubrg VA, 24061, USA. 
The work is supported by the US National Science Foundation under grants ECCS-1811497 and CCF-1937487. The corresponding author is L. Liu (ljliu@ieee.org).}
}

\maketitle

\IEEEpubidadjcol

\begin{abstract}
Deep reinforcement learning (DRL) has been shown to be successful in many application domains.
Combining recurrent neural networks (RNNs) and DRL further enables DRL to be applicable in non-Markovian environments by capturing temporal information. 
However, training of both DRL and RNNs is known to be challenging requiring a large amount of training data to achieve convergence.
In many targeted applications, such as those used in the fifth generation (5G) cellular communication, the environment is highly dynamic while the available training data is very limited. 
Therefore, it is extremely important to develop DRL strategies that are capable of capturing the temporal correlation of the dynamic environment requiring limited training overhead. 
In this paper, we introduce the deep echo state Q-network (DEQN) that can adapt to the highly dynamic environment in a short period of time with limited training data. 
We evaluate the performance of the introduced DEQN method under the dynamic spectrum sharing (DSS) scenario, which is a promising technology in 5G and future 6G networks to increase the spectrum utilization. 
Compared to conventional spectrum management policy that grants a fixed spectrum band to a single system for exclusive access, DSS allows the secondary system to share the spectrum with the primary system.
Our work sheds light on the application of an efficient DRL framework in highly dynamic environments with limited available training data.
\end{abstract}


\section{Introduction}
In the last few years, deep reinforcement learning (DRL) has been widely adopted in different fields, ranging from playing video games~\cite{mnih2013DRL}, playing chess~\cite{silver2016Go}, to robotics~\cite{levine2016Robotic}.
DRL provides a flexible solution for many types of problems due to the fact that it does not need to model complex systems or to label data for training.
Utilizing recurrent neural networks (RNNs) in DRL, the deep recurrent Q-network (DRQN) is introduced to process the temporal correlation of input sequences in a non-Markovian environment~\cite{hausknecht2015DRQN}.
Even though DRQN is a powerful machine learning tool, it faces serious issues related to training due to the following two reasons: 1) DRL requires a relatively large amount of training data and computational resources to make the learning agent converge to an appropriate policy, which is a major bottleneck for applying DRL to many real-world applications~\cite{he2017learning}.
2) The kernel of DRQN, the RNN, has issues related to vanishing and exploding gradients that make the underlying training difficult~\cite{2013RNNdifficulty}.
Therefore, the difficulties of training DRL agents and RNNs make the training of DRQNs an extremely challenging problem and prevent it from being widely adopted for analyzing time-dynamic applications.
In light of the training challenges, in this work we exploit a special type of RNNs, echo state networks (ESNs), to reduce the training time and the required training data~\cite{jaeger2001echo}.
ESNs simplify the underlying RNNs training by only training the output weights while leaving input weights and recurrent weights untrained.
Existing research shows that ESNs can achieve comparable performance with RNNs, especially in some tasks requiring fast learning~\cite{tanaka2019RC}.
Accordingly, in this work, we adopt ESNs as the Q-networks in the DRL framework, which is referred to as deep echo state Q-networks (DEQN).
We will show that DEQN has the benefit of learning a good policy with short training time and limited training data.

Fueled by the popularity of smartphones as well as the upcoming deployment of the fifth generation (5G) mobile broadband networks, mobile data traffic will grow at a compound annual growth rate (CAGR) of 46 percent between 2017 and 2022, reaching 77.5 exabytes (EB) per month by 2022~\cite{cisco2018VNI}. 
A significant portion of these data traffic will be real-time or delay-sensitive. 
For example, live video will grow 9-fold from 2017 to 2022 while virtual reality and augmented reality traffic will increase 12-fold at a CAGR of 63 percent. 
This suggests that future wireless networks will likely face the pressing demand of being able to conduct real-time processing for large volume data in an efficient way. 
In 5G networks, massive connectivity is regarded as a primary use case with dynamic spectrum sharing (DSS) as an enabling technology.
In fact, DSS has been announced as the key technology for 5G by many companies and operators around the world including Qualcomm, Ericsson, AT\&T, and Verizon~\cite{DSS19,VerizonDSA19}.
Unlike the current static spectrum management policy that gives a single system exclusive right to access the spectrum, DSS has a more flexible policy by adopting a hierarchical access structure with primary users (PUs) and secondary users (SUs)~\cite{ahmad20205g}.
SUs are allowed to access the licensed spectrum when PUs receive tolerable interference.

Obtaining control information from the environment is costly in 5G mobile wireless networks.
First, a SU cannot detect the activities of all PUs simultaneously because performing spectrum sensing is energy-consuming.
Second, exchanging control information between wireless devices imposes a control overhead in wireless network operations.
Therefore, the major challenge of DSS is how to optimize the system performance under limited information exchange between the secondary system and the primary system.
DRL is a suitable framework for developing DSS strategies because of its abilities to adapt to unknown environment without modeling the complex 5G networks.
DRL usually requires tons of training data and long training time.
However, wireless networks are dynamic due to factors such as path loss, shadow fading, and multi-path fading~\cite{tse2005fundamentals}, which largely decreases the number of effective training data that reflect the latest environment.
Furthermore, the performance of spectrum sharing depends on access strategies of multiple users.
If one user changes its access strategy, then other users have to change their access strategies accordingly.
Under these circumstances, the number of effective training samples reflected in the latest wireless environment will be extremely limited. 
As a result, designing an efficient DRL framework only requiring a small amount of training data will be critical for 5G and future 6G DSS networks.
In this work, we introduce DEQN to learn a spectrum access strategy for each SU in a distributed fashion with limited training data and short training time in the highly dynamic 5G networks.

The main contributions of our work are as follows:
\begin{itemize}
\item We design an efficient DRL framework, DEQN, to adapt to highly dynamic environment with limited training data and provide training strategies for the introduced DEQN.
\item We apply the DEQN method in the critical problem of DSS for 5G networks where the system is highly dynamic and interactive. Compared to existing DRL-based strategies, our method can quickly adapt to real mobile wireless environment to achieve improved network performance under limited training data.
\item This work is the first to formulate a DRL strategy that jointly considers spectrum sensing and spectrum sharing in the underlying DSS network for 5G.
\end{itemize}

\section{Problem Definition for DSS}
In this section, we introduce the DSS problem and discuss its challenges.
We consider a DSS system where the primary network consisting of $M$ PUs and the secondary network consisting of $N$ SUs.
It is assumed that one wireless channel is allocated to each PU individually and cross-channel interference is negligible.
We consider a discrete time model, where the dynamics of the DSS system, such as behaviors of users and changes of the wireless environment, are constrained to happen at discrete time slots $t$ ($t$ is a natural number).
Our goal is to develop a distributive DSS strategy for each SU to increase the spectrum utilization without harming the primary network's performance.

\begin{figure}[t]
\centering
\includegraphics[width=0.9\linewidth]{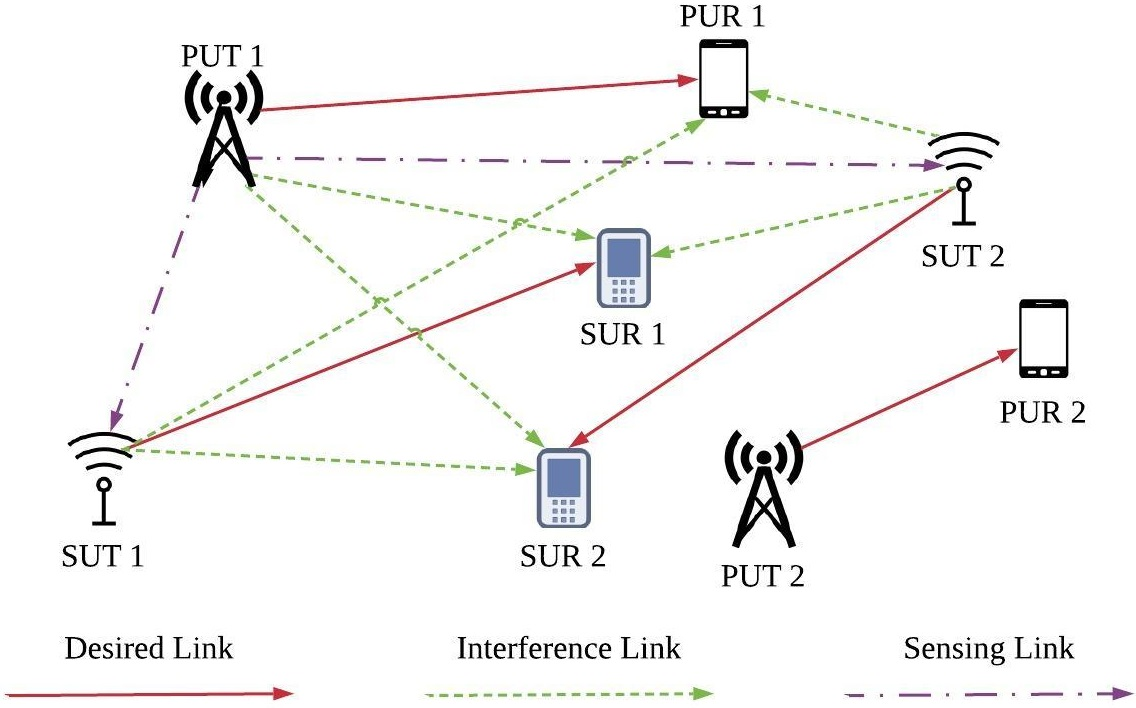}
\caption{The desired links, the interference links, and the sensing links when PU1, SU1, and SU2 are operating on the same channel. PUT/SUT represent the transmitters of PU/SU and PUR/SUR represent the receivers of PU/SU. }
\label{fig:system_model}
\end{figure}

The data of an user are transmitted over the wireless link between its transmitter and receiver.
Signal-to-interference-plus-noise ratio (SINR) is a quality measure of the wireless connection that compares the power of a desired signal to the sum of the interference power and the power of background noise.
The higher value of the SINR, the better quality of the wireless connection.
The SINR of the user $k$'s wireless connection on channel $m$ at time slot $t$ is written as
\begin{equation}
    \text{SINR}^{k}_{m} [t] = \frac{P^k \cdot \left| H^k[t] \right|^2}{\sum\limits_{z \in \Phi_m^k} P^z \cdot \left| H^{zk}[t] \right|^2 + N_m}
\label{eqn:sinr}
\end{equation}
where $P^k$ and $P^z$ are the transmit power of the user $k$ and the user $z$, respectively, $\Phi_m^k$ is the set containing all the users that are transmitting on channel $m$ except for the user $k$, $H^k[t]$ is the channel gain of the desired link of the user $k$, $H^{zk}[t]$ is the channel gain of the interference link between the user $z$'s transmitter and the user $k$'s receiver, and $N_m$ is the background noise power on channel $m$.
Note that all channel gains are changing over time so SINR is also time-variant.
The desired link is the link between the transmitter and the receiver of the same user.
The interference link is the link between the transmitter and the receiver of two different users if these two users are transmitting on the same channel simultaneously.
Figure~\ref{fig:system_model} shows the complicated association of desired links and interference links when PU1, SU1, and SU2 are operating on the same channel.
Since cross-channel interference is negligible, the interference link between two users operating on different channels is out of consideration.
The radio signal attenuates as it propagates through space between the transmitter and the receiver, which is referred to as the path loss.
In addition to the path loss, the channel gain is affected by many factors such as shadow fading and multi-path fading.
Shadow fading is caused by a large obstacle like a hill or a building obscuring the main signal path between the transmitter and the receiver.
Multi-path fading occurs in any environment where multiple propagation paths exist between the transmitter and the receiver, which may be caused by reflection, diffraction, or scattering.
In telecommunication society, the channel model is carefully designed to be consistent with wireless field measurements.
We generate channel gains based on the WINNER II channel model~\cite{winner2}, which is widely used in industry to make fair comparisons of telecommunication algorithms.

To enable the protection of the primary network, we assume that a PU will broadcast a warning signal if its data transmission experiences a low SINR.
There are two possible causes for low SINR.
First, the wireless connection of the desired link of the PU is in deep fade, which means the channel gain of the desired link is low.
This leads to a small value of the numerator in Equation (\ref{eqn:sinr}) so SINR is low.
Second, the signals from one or more SUs cause strong interference to a PU when they are transmitting over the same wireless channel at the same time. 
This leads to a large value of the denominator in Equation (\ref{eqn:sinr}), so SINR assumes a low value again.
We called SUs "collides" with the PU in this case.
The warning signal contains information related to which PU may be interfered so that the SUs transmitting on the same channel are aware of the issue. 
In fact, this kind of warning signal is similar to the control signals (e.g. synchronization, downlink/uplink control) used in current 4G and 5G networks.
It is common to assume that the control signals are received perfectly at receivers, otherwise the underlying network will not even work.
In reality, the control signal can be transmitted through a dedicated control channel.
According to this mechanism, a PU will broadcast a warning signal once the received SINR is low, and this is the only control information from the primary system to the secondary system to enable the protection for PUs under DSS.
Note that a PU may send a warning signal even when no collisions happen because of deep fade.

The activity of a PU consists of two states: (1) \textit{Active} and (2) \textit{Inactive}.
If a PU is transmitting data, it is in \textit{Active} state, otherwise it is in \textit{Inactive} state.
A spectrum opportunity on a channel occurs when the licensed PU of that channel is in \textit{Inactive} state or any SU can transmit on that channel with little interference to the \textit{Active} licensed PU.
Unfortunately, it is difficult for a SU to obtain the information of activity states of PUs or the interference that it will cause in the highly dynamic 5G networks.
A SU has to perform spectrum sensing to detect the activity of a PU, but the accuracy of detection is based on the wireless link between the transmitters of the PU and the SU, the background noise, and the transmit power of the PU.
On the other hand, the interference level caused by a SU is determined by the interference link from the SU to the PU, the desired link of the PU, transmit powers of the PU and the SU, and the background noise.
Furthermore, all these factors for determining spectrum opportunities are time-variant so control information becomes outdated quickly.
Since obtaining control information is costly in 5G mobile wireless networks, it is impractical to design a DSS strategy by assuming that all the control information is known.

\begin{figure}[ht]
\centering
	\includegraphics[width=1.0\linewidth]{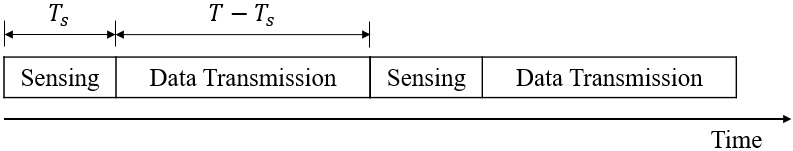}
	\caption{The time structure of spectrum sensing and data transmission.}
	\label{fig:sensing_period}
\end{figure}

SUs should provide protection to prevent PUs from harmful interference since the primary system is the spectrum licensee.
A commonly used method is that the transmitter of a SU performs spectrum sensing to detect the activity of a PU before accessing a channel.
Due to the power and complexity constraints, a SU is unable to perform spectrum sensing across all channels simultaneously.
Therefore, we assume that a SU can only sense one channel at a particular time.
We adopt the energy detector as the underlying spectrum sensing method, which is the most common one due to its low complexity and cost.
The energy detector of SU $n$ first computes the energy of received signals on channel $m$ as follows:
\begin{equation}
E_m^n [t] = \sum_{t'=t}^{t+T_s-1} \left| y_m^n[t'] \right|^2
\label{eqn:sensed_energy}
\end{equation}
where $t$ is the starting time slot of the spectrum sensing, $y_m^n[t']$ is the received signal at time slot $t'$, and $T_s$ is the number of time slots of the spectrum sensing.
We consider the half-duplex SU system where a SU cannot transmit data and perform spectrum sensing at the same time.
We assume a periodic time structure of spectrum sensing and data transmission as shown in Figure~\ref{fig:sensing_period}.
%
To be specific, the $k^{th}$ sensing and transmission period contains $T$ time slots from $kT + 1$ to $(k+1)T$, the spectrum sensing contains the first $T_s$ time slots in the period from $kT + 1$ to $kT + T_s$, and the data transmission contains the subsequent $T-T_s$ time slots in the period from $kT + T_s + 1$ to $(k+1)T$.

The received signal $y_m^n[t']$ depends on the activity state of PU $m$, the power of PU $m$, the background noise, and the sensing link between the transmitters of PU $m$ and SU $n$.
When PU $m$ is in the \textit{Inactive} state, the received signal is represented as
\begin{equation}
y_m^n[t'] = \omega_m[t']
\end{equation}
When PU $m$ is in the \textit{Active} state, the received signal is represented as
\begin{equation}
y_m^n[t'] = \sqrt{P^m} \cdot H^{mn}[t'] + \omega_m[t']
\end{equation}
where $\omega_m[t'] \sim \mathcal{CN}(0, N_m)$ is a circularly-symmetric Gaussian noise with zero mean and variance $N_m$, $P^m$ is the transmit power of PU $m$, and $H^{mn}[t]$ is the channel gain of the sensing link between the transmitters of PU $m$ and SU $n$.
If the energy computed in Equation (\ref{eqn:sensed_energy}) is higher than a threshold, the PU is considered in the \textit{Active} state, otherwise the PU is considered in the \textit{Inactive} state.
The challenge of designing an energy detector is how to set the threshold properly.
The value of the threshold is actually a trade-off between the detection probability and the false alarm probability.
However, setting the threshold for achieving a good trade-off is related to many factors, including the channel gain of the sensing link, the transmit power of the PU, the noise variance, the number of received signals, etc. 
This information is difficult to obtain before deploying in the real environment and is time-variant.
Furthermore, setting a threshold is difficult in some cases because of the relative positions of transmitters and receivers.
As shown in Figure~\ref{fig:system_model}, the sensing link is between the transmitters of the PU and the SU, but the interference link is between the transmitter of the SU and the receiver of the PU.
The discrepancy between the sensing link and the interference link may cause the hidden node problem, where the sensing link is weak but the interference link is strong.
For example, the transmitters of a SU and a PU are far away from each other while the SU transmitter is close to the receiver of the PU.
In this case, the transmitters of the SU and the PU are hidden nodes with respect to each other.
%
%
%
%
The warning signals from PUs are designed to provide additional protection to the primary system for the case where the SU cannot detect the activity of the PU, thereby mitigating the issues caused by the hidden nodes. 
Meanwhile, instead of making the spectrum access decision solely based on the outcomes of the energy detector, we developed a DRL framework to construct a novel spectrum access policy: The DRL agent will use the sensed energy as the input to learn a spectrum access strategy to maximize the cumulative reward.
The reward is designed to maximize the spectral-efficiencies of SUs while enabling the protection for PUs with the help of warning signals from PUs.

\section{DRL Framework for DSS and DEQN}

\subsection{Background on DRL}
\label{subsec:background_DRL}
RL is one type of machine learning method that provides a flexible architecture for solving many types of practical problems because it does not need to model complex systems or to label data for training.
In RL, an agent learns how to select actions to maximize the cumulative reward in a stochastic environment.
The dynamics of the environment is usually modeled as a Markov decision process (MDP), which characterized by a tuple $(\mathcal{S}, \mathcal{A}, \mathcal{P}, R, \mathcal{\gamma})$, where $\mathcal{S}$ is the state space, $\mathcal{A}$ is the action space, $\mathcal{P}$ is the state transition providing $\Pr(s_{t+1} |s_{t}, a_{t})$, $R$ is the reward function providing $r_t = R(s_{t}, a_{t})$, and $\gamma$ is a discount factor for calculating cumulative reward.
Specifically, at time $t$, the state is $s_{t} \in \mathcal{S}$, the RL agent selects an action $a_{t} \in \mathcal{A}$ by following a policy $\pi(s_t)$ and receives the reward $r_t$, and then the system shifts to the next state $s_{t+1}$ according to the state transition probability.
Note that the action $a_{t}$ affects both the immediate reward $r_t$ and the next state $s_{t+1}$.
Consequently, all subsequent rewards are affected by the current action.
The goal of RL agent is to find a policy $\pi$ to maximize the cumulative reward, $\mathbb{E}_{\pi} \left[ \sum_{t=1}^{\infty} \gamma^{t-1} r_{t} \right]$.

In RL, a model-free algorithm does not require state transition probability for learning, which is useful when the underlying system is complicated and difficult to model.
Q-learning~\cite{watkins1992qlearn} is the most widely used model-free RL algorithm that aims to find the Q-function of each state-action pair for a given policy, which is defined as
\begin{equation}
    Q^{\pi}(s_t, a_t) = \mathbb{E} \left[ \sum_{t' = 1}^{\infty} \gamma^{t'-1} r_{t'}  \mid s_1 = s_t, a_1 = a_t  \right].
\end{equation}
Q-function represents the cumulative reward when taking action $a_t$ in the state $s_t$ and then following policy $\pi$.
Q-learning constructs a Q-table to estimate the Q-function of each state-action pair by iteratively updating each element of the Q-table through dynamic programming.
The update rule of the Q-table is given as follows:
\begin{equation}
\begin{aligned}
    Q(s_t, a_t) & \leftarrow  Q(s_t, a_t)  \\
    &+ \alpha \left[ r_t + \gamma \max\limits_{a} Q(s_{t+1}, a) - Q(s_t, a_t) \right],
\end{aligned}
\end{equation}
where $\alpha \in (0,1)$ is the learning rate. 
The policy $\pi$ that selects action is the $\epsilon$-greedy policy as follows:
\begin{equation}
a_{t} =
    \begin{cases}
    \argmax_a Q(s_t, a) & \text{, with probability } 1-\epsilon, \\
    \text{random action} & \text{, with probability }\epsilon,
    \end{cases}
\end{equation}
where $\epsilon \in [0, 1]$ is the exploration probability.
However, Q-learning performs poorly when the dimension of the state is high because updating a large Q-table makes training difficult or even impossible.

Deep Q-Networks (DQN)~\cite{mnih2013DRL} is introduced to solve high-dimensional state problems by leveraging a neural network as the function approximator of the Q-table,  which is referred to as the Q-network.
Specifically, the Q-network takes the state $s_t$ as input and outputs the estimated Q-function of all possible actions.
One key approach of DQN to improve the training stability is by creating two Q-networks: the evaluation network $Q(s, a; \theta)$ and the target network $Q(s, a; \theta^{-})$.
The target network is used to generate the targets for training the evaluation network while the evaluation network is used to determine the actions.
The loss function for training the evaluation network is written as 
\begin{equation}
    \left( r_t + \gamma \max\limits_{a} Q(s_{t+1}, a; \theta^{-}) - Q(s_t, a_t; \theta) \right)^2,
\end{equation}
where $r_t + \gamma \max\limits_{a} Q(s_{t+1}, a; \theta^{-})$ is the target Q-value.
The weights of the target network $\theta^{-}$ is periodically synchronized with the weights of the evaluation network $\theta$.
The purpose is to fix targets temporarily during training to improve the training stability of the evaluation network.

An improvement of DQN to prevent overestimation of Q-values is called double Q-learning~\cite{van2016ddqn}, where the evaluation network is used to select the action when computing the target Q-value, but the target Q-value is still generated by the target network.
Specifically, the target Q-value for the evaluation network is calculated by
\begin{equation}
    r_t + \gamma  Q(s_{t+1}, a'; \theta^{-}),
\end{equation}
where $a' = \argmax\limits_{a} Q(s_{t+1}, a; \theta)$.
Double Q-learning can improve the accuracy in estimating Q-function, thereby improves the learned policy.

\subsection{Existing DRL-based Strategies for DSS}
DRL-based methods have recently been applied in dynamic spectrum access (DSA) networks~\cite{wang2018deep,naparstek2018deep, Chang2019DSA} where the focus is exclusively on the "access" part of the problem with over-simplified network setup.
To be specific, \cite{wang2018deep} considers single SU selects one channel to access in the multichannel environment, and the goal is to maximize the number of selecting good channels for access.
\cite{naparstek2018deep} assumes that the available spectrum channels are known a priori and develops a centralized spectrum access algorithm for multi-user access.
Both \cite{wang2018deep} and \cite{naparstek2018deep} assumes that one channel can only be used by one user at any particular time.
Although \cite{Chang2019DSA} considers multiple SUs can access a channel at the same time, a SU cannot access a channel that a PU is using.
\cite{Chang2019DSA} also assumes that each SU can sense all channels simultaneously and the collision between a PU and a SU can be perfectly detected.
In this work, in order to provide a comprehensive study for the impact of DEQN on relevant DSS networks for 5G, we consider practical situations of DSS where 1) mobile users cannot conduct spectrum sensing perfectly. 2) mobile users cannot sense multiple channel at a particular time. 3) there are multiple PUs and multiple SUs in a DSS network. 4) A channel can be shared by multiple users if the interference between them is weak.
Furthermore, unlike previous work which utilizes binary ACK/NACK feedback as the reward function, we calculate the practical reward based on the spectral-efficiency of each mobile link. 
To be closely in line with the real wireless environment, the spectral-efficiency of a mobile link is calculated using the transmission procedure defined in the telecommunication standard.
In this way, we can train and evaluate the underlying DEQN-based DRL strategies in realistic 5G application scenarios.
It is important to note that in our work we treat the unprocessed soft spectrum sensing information as the input states of the DRL agent.
Soft spectrum sensing information can be directly obtained from spectrum sensing sensors. 
Through the soft spectrum sensing input, the DRL agent will learn an appropriate detection criterion for each SU that adapts to different mobile wireless environments, geometry of mobile users, and activities of mobile users.
This is indeed the first work to study DSS that combines soft spectrum sensing information and spectrum access strategies through the DRL framework.

\subsection{DRL Problem Formulation for DSS}

%
%
%
%
%
%
%

We now formulate the DSS problem using the DRL framework, where all SUs in the secondary system learn their spectrum access strategies in a distributed fashion through the interactions with the mobile wireless environment.
To be specific, we assume that each SU has a DRL agent that takes its observed state as the input and learns how to perform spectrum sensing and access actions in order to maximize its cumulative reward.
The reward for each SU is designed to maximize its spectrum efficiency and to prevent harmful interference to PUs.

The state of SU $n$ in the $k^{\text{th}}$ sensing and transmission period is denoted by 
\begin{equation}
s^n[k] = \left(E^n [k], Q^n [k] \right),
\label{eqn:state}
\end{equation}
where $k$ is a non-negative integer, $E^n[k]$ is the energy of received signals, and $Q^n[k]$ is a one-hot $M$-dimensional vector indicating the sensed channel from time slots $kT + 1$ to $kT + T_s$.
If the index of the sensed channel is $m$, then the $m^{\text{th}}$ element of $Q^n[k]$ is equal to one while other elements of $Q^n[k]$ are zeros.
On the other hand, $E^n [k]$ is equal to $E^n_m [kT]$ that is calculated by Equation (\ref{eqn:sensed_energy}).
The action of SU $n$ in the $k^{\text{th}}$ sensing and transmission period is denoted by
\begin{equation}
a^n[k] = \left( q^n[k], z^n[k] \right),
\end{equation}
where $q^n[k] \in \{0, 1\}$ represents SU $n$ will either access the current sensed channel ($q^n[k] = 1$) or be idle ($q^n[k] = 0$) during the data transmission part of the $k^{\text{th}}$ period (from time slots $kT + T_s + 1$ to $(k+1)T$), $z^n[k] \in \{1, ..., M\}$ represents SU $n$ will sense channel $z^n[k]$ during the sensing part of the $(k+1)^{\text{th}}$ period (from time slots $(k+1)T + 1$ to $(k+1)T + T_s$).
In other words, SU $n$ makes two decisions: $q^n[k]$ decides whether to conduct data transmission in the current sensed channel of the $k^{\text{th}}$ period and $z^n[k]$ decides which channel to sense in the $(k+1)^{\text{th}}$ period.
Therefore, the dimension of each SU's action space is $2M$.
Note that the sensed channel in the $k^{\text{th}}$ period may be different from that in the $(k+1)^{\text{th}}$ period

\begin{table}[ht]
\centering
\caption{The SINR and CQI mapping to modulation and coding rate.
}
\label{tab:cqi_table} 
\begin{tabular}{ccccc}
\toprule
CQI index & SINR & modulation & code rate & efficiency \\
& ($\geq$) &  & ($\times$1024) & (bits per symbol)\\
\midrule
0  & \multicolumn{4}{c}{out of range}
 \\
 1  & -6.9360  & QPSK &   78 & 0.1523
 \\
 2  & -5.1470  & QPSK &  120 & 0.2344
 \\
 3  & -3.1800  & QPSK &  193 & 0.3770
 \\
 4  & -1.2530  & QPSK &  308 & 0.6016
 \\
 5  & 0.7610   & QPSK &  449 & 0.8770
 \\
 6  & 2.6990   & QPSK &  602 & 1.1758
 \\
 7  & 4.6940   & 16QAM & 378 & 1.4766
 \\
 8  & 6.5250   & 16QAM & 490 & 1.9141
 \\
 9  & 8.5730   & 16QAM & 616 & 2.4063
 \\
 10 & 10.3660  & 64QAM & 466 & 2.7305
 \\
 11 & 12.2890  & 64QAM & 567 & 3.3223
 \\
 12 & 14.1730  & 64QAM & 666 & 3.9023
 \\
 13 & 15.8880  & 64QAM & 772 & 4.5234
 \\
 14 & 17.8140  & 64QAM & 873 & 5.1152
 \\
 15 & 19.8290  & 64QAM & 948 & 5.5547
 \\
\bottomrule
\end{tabular}
\end{table}

In our work, we use a discrete reward function which is similar to the existing DRL-based DSS methods. 
Compared to a simple binary reward ($0$ and $+1$, $-1$ and $+1$) in \cite{wang2018deep} and \cite{naparstek2018deep}, we consider a more relevant and comprehensive reward design that is based on the underlying achieved modulation and coding strategy (MCS) adopted in the 3GPP LTE/LTE-Advanced standard~\cite{LTE}.
To be specific, a receiver measures SINR to evaluate the quality of the wireless connection and feedback the corresponding Channel Quality Indicator (CQI) to the transmitter~\cite{liu2012downlink}.
In this work, we follow the method presented in~\cite{chiumento2015CQI} to map the received SINR to the CQI.
After receiving the CQI, the transmitter determines the MCS for data transmission based on the CQI table specified in the 3GPP standard~\cite{LTE}.
The SINR and CQI mapping to MCS is given in Table~\ref{tab:cqi_table} for reference.
Accordingly, the achieved spectral-efficiency can be calculated by (bits/symbol) = (modulation’s power of 2) $\times$ (code rate) representing the average information bits per symbol. 
This critical metric is utilized as the reward function of our design.

To jointly consider the performance of the primary and the secondary systems, the reward function corresponding to SU $n$ accessing channel $m$ depends on both the spectral-efficiency of SU $n$ and PU $m$.
During time slots $kT + T_s + 1$ to $(k+1)T$, the average spectral-efficiency of SU $n$, $\bar{e}^n[k]$, and the average spectral-efficiency of PU $m$, $\bar{e}^m[k]$, are calculated by
\begin{align}
\begin{aligned}
\bar{e}^n[k] &= \frac{1}{T-T_s} \sum_{t'=kT+T_s}^{(k+1)T-1} e_m^n[t'] \\
\bar{e}^m[k] &= \frac{1}{T-T_s} \sum_{t'=kT+T_s}^{(k+1)T-1} e_m^m[t']
\end{aligned}
\label{eqn:efficiency_SU}
\end{align}
where $e_m^n[t']$ and $e_m^m[t']$ represent the spectral-efficiency of SU $n$ and PU $m$ on channel $m$ at time slot $t'$, respectively.
The reward of SU $n$ in the $k^{\text{th}}$ transmission period is defined as
\begin{equation}
{r^n[k]} = 
\begin{cases}
-2, & \text{if } \bar{e}^m[k] < 1.5 \\
-1, & \text{if SU $n$ is idle in the $k^{\text{th}}$ period}\\
0, & 
\text{if } \bar{e}^m[k] \geq 1.5 \text{ and } \bar{e}^n[k] < 1 \\
1, & 
\text{if } \bar{e}^m[k] \geq 1.5 \text{ and } 1 \leq \bar{e}^n[k] < 2 \\
2, & 
\text{if } \bar{e}^m[k] \geq 1.5 \text{ and } 2 \leq \bar{e}^n[k] < 3 \\
3, & 
\text{if } \bar{e}^m[k] \geq 1.5 \text{ and } \bar{e}^n[k] \geq 3 \\
\end{cases}
\label{eqn:reward_def}
\end{equation}
To enable the protection for the primary system, PU $m$ will broadcast a warning signal if its average spectral-efficiency is below $1.5$, and then the reward received by SU $n$ that accesses channel $m$ is set to $-2$. 
To motivate SUs to explore spectrum opportunities, the reward $r^n[k]$ is set to $-1$ if SU $n$ decides to be idle in the $k^{\text{th}}$ transmission period.
When PU $m$ does not suffer from strong interference (the average spectral-efficiency of PU $m$ is larger than $1.5$), we increase the reward $r^n[k]$ from $0$ to $3$ as the average spectral-efficiency of SU $n$ increases (see Equation (\ref{eqn:reward_def})).
Note that the low spectral-efficiency of a PU or a SU does not necessarily mean collisions because the underlying wireless channels are changing dynamically over time.
If the channel gain of the wireless link is small, the spectral-efficiency of the user will be low even if there is no collision.
Therefore, the reward function and the warning signal are introduced since it is impossible to detect collisions perfectly in practical wireless environments.

\subsection{Efficient Training for DEQN}
\begin{figure}[ht]
\centering
	\includegraphics[width=.9\linewidth]{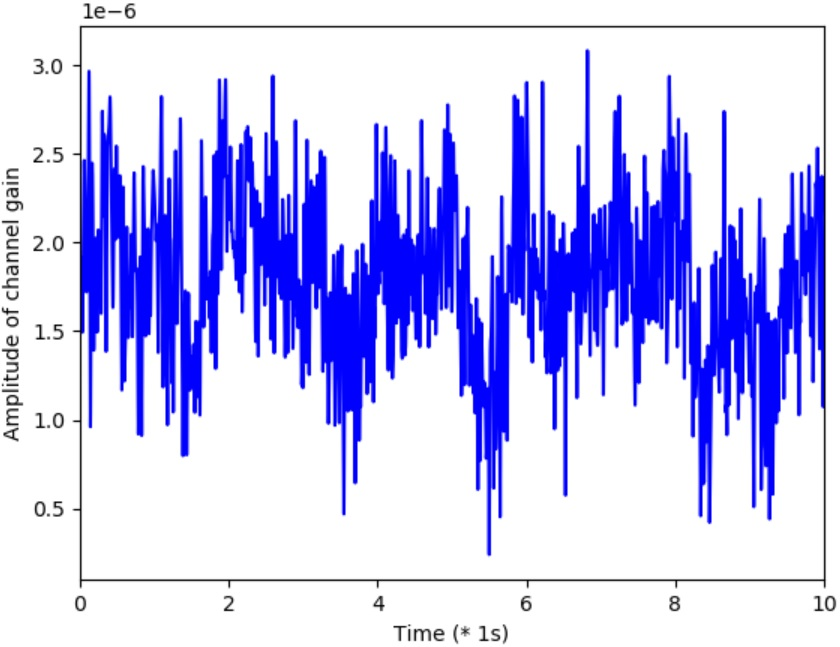}
	\caption{Time-variant channel gain of a wireless link.}
	\label{fig:wireless_channel}
\end{figure}

To capture the activity patterns of PUs, which are usually time-dependent, applying DRQNs is a natural choice.
Although DQNs are able to learn the temporal correlation by stacking a history of states in the input, the sufficient number of stacked states is unknown because it depends on PUs' behavior patterns.
RNNs are a family of neural networks for processing sequential data without specifying the length of temporal correlation.
However, the training of RNNs is known to be difficult that suffers from vanishing and the exploding gradients problems.
Furthermore, the required amount of training data for achieving convergence is large in the DRL scheme, since there are no explicit labels to guide the training and the agents have to learn from interacting with its environment.
In the wireless environment, the channel gain of a wireless link changes rapidly, which is shown in Figure~\ref{fig:wireless_channel}.
Note that the environment observed by a SU is affected by other SUs' access strategies because of possible collisions between SUs, and all SUs are dynamically adjusting their DSS strategies during their training processes.
As a result, in the DSS problem, the duration for a learning environment being stable is short and the available training data is very limited.

\begin{algorithm*}[t]
\caption{The training algorithm for DEQN.}
\begin{algorithmic}
\STATE {
Initialize the wireless environment with $M$ PUs and $N$ SUs.
}
\STATE {
Set the sensing and transmission period to $T$ time slots and the sensing duration to $T_s$ time slots.
}
\STATE {
Set the buffer size to $Z$, the training iteration to $I$, and the exploration probability to $\epsilon$.
}
\STATE {
Randomly initialize an evaluation network $\text{DEQN}^n_{\theta}$ and a target network $\text{DEQN}^n_{\theta^{-}}$ with the same weights for each SU $n$.
}
\STATE {
Each SU $n$ randomly selects one channel ($= z^n[0]$) to sense for $T_s$ time slots and then computes the state $s^n[1]$.
}
\FOR {$q = 1, ...$}
\STATE {
Initialize an empty buffer $B^n_q$ for each SU.
}
\FOR {$z = 1, ..., Z$}
\STATE {
Let $k = (q-1)Z + z$.
}
\STATE {
Each SU $n$ inputs $s^n[k]$ to $\text{DEQN}^n_{\theta}$, calculates the hidden state $h^n_{\theta}[k]$, and outputs $o^n_{\theta}[k]$.
}
\STATE {
Each SU $n$ decides action $a^n[k] = \left( q^n[k], z^n[k] \right)$ based on $\epsilon$-greedy policy, where $a^n[k]$ is the index of the maximum element of $o^n_{\theta}[k]$ with probability $1 - \epsilon$ and $a^n[k]$ is chosen randomly with probability $\epsilon$.
}
\STATE {
Each SU $n$ accesses channel $z^n[k-1]$ if $q^n[k]=1$ or does not access if $q^n[k]=0$ for $T-T_s$ time slots.
}
\STATE {
Each SU $n$ obtains the reward $r^n[k]$ according to Equation (\ref{eqn:reward_def}).
}
\STATE {
Each SU $n$ senses channel $z^n[k]$ for $T_s$ time slots and then computes the state $s^n[k+1]$.
}
\STATE {
Each SU $n$ inputs $s^n[k+1]$ to $\text{DEQN}^n_{\theta^{-}}$, calculates the hidden state $h^n_{\theta^{-}}[k]$, and outputs $o^n_{\theta^{-}}[k]$.
}
\STATE {
Each SU $n$ stores $(s^n[k], h^n_{\theta}[k], a^n[k], r^n[k], s^n[k+1], h^n_{\theta^{-}}[k])$ in $B^n_q$.
}
\ENDFOR
\FOR {iteration $= 1, ..., I$}
\STATE {
Each SU $n$ samples random training batch $(s^n[k], h^n_{\theta}[k], a^n[k], r^n[k], s^n[k+1], h^n_{\theta^{-}}[k])$ from $B^n_q$.
}
\STATE {
Each SU $n$ inputs $s^n[k]$ and $h^n_{\theta}[k]$ to $\text{DEQN}^n_{\theta}$ to calculate $o^n_{\theta}[k]$
}
\STATE {
Each SU $n$ inputs $s^n[k+1]$ and $h^n_{\theta}[k+1]$ to $\text{DEQN}^n_{\theta^{-}}$ to calculate $o^n_{\theta^{-}}[k]$
}
\STATE {
Each SU $n$ updates $\text{DEQN}^n_{\theta}$ by performing gradient descent step on $\left( r^n[k] + \gamma o^n_{y, \theta^{-}}[k+1] - o^n_{y, \theta}[k] \right)^2$, where $y$ is the index of the maximum element of $o^n_{\theta}[k+1]$.
}
\ENDFOR
\STATE {
Each SU $n$ synchronizes $\text{DEQN}^n_{\theta^{-}}$ with $\text{DEQN}^n_{\theta}$.
}
\ENDFOR
\end{algorithmic}
\label{alg:DEQN_training}
\end{algorithm*}

The standard training technique for RNNs is to unfold the network in time into a computational graph that has a repetitive structure, which is called backpropagation through time (BPTT).
BPTT suffers from the slow convergence rate and needs many training examples.
DRQN also requires a large amount of training data because a learning agent finds a good policy by exploring the environment with different potential policies.
Unfortunately, in the DSS problem, there are only limited training data for a stable environment due to dynamic channel gains, partial sensing, and the existence of multiple SUs.
To address this issue, we use ESNs as the Q-networks in the DRQN framework to rapidly adapt to the environment.
ESNs simplify the training of RNNs significantly by keeping the input weights and recurrent weights fixed and only training the output weights.
%

%
We denote the sequence of states for SU $n$ by $\{ s^n[1], s^n[2], ...\}$. 
Accordingly, the sequence of hidden states, $\{ h^n[1], h^n[2], ...\}$, is updated by
\begin{equation}
\begin{aligned}
h^n[k] = & (1-\beta) \cdot h^n[k-1] \\
&+ \beta \cdot \text{tanh} \left( W^n_{in} s^n[k] + W^n_{rec} h^n[k-1] \right), 
\end{aligned}
\label{eqn:hidden_state}
\end{equation}
where $W^n_{in}$ is the input weight, $W^n_{rec}$ is the recurrent weight, $\beta \in [0,1]$ is the leaky parameter, and we let $h^n[0] = \boldsymbol{0}$.
The output sequence, $\{o^n[1], o^n[2], ...\}$, is computed by
\begin{equation}
o^n[k] = W^n_{out} u^n[k]
\end{equation}
where $u^n[k]$ is a concatenated vector of $s^n[k]$ and $h^n[k]$, and $W^n_{out}$ is the output weight.
Note that the output vector $o^n[k]$ is a $2M$-dimensional vector, where each element of $o^n[k]$ corresponds to the estimated Q-value of selecting one of all possible actions given the state $s^n[1], ..., s^n[k]$.

The double Q-learning algorithm~\cite{van2016ddqn} is adopted to train the underlying DEQN agent of each SU.
As discussed in Section \ref{subsec:background_DRL}, each DEQN agent has two Q-networks: the evaluation network and the target network.
Let the output sequence from the evaluation network and the target network be $\{o^n_{\theta}[1], o^n_{\theta}[2], ...\}$ and $\{o^n_{\theta^{-}}[1], o^n_{\theta^{-}}[2], ...\}$, respectively.
The loss function for training the evaluation network of SU $n$ is written as
\begin{equation}
    \left( r^n[k] + \gamma o^n_{y, \theta^{-}}[k+1] - o^n_{y, \theta}[k] \right)^2,
\end{equation}
where $o^n_{y, \theta^{-}}[k+1]$ and $o^n_{y, \theta}[k]$ are the $y^{\text{th}}$ element of $o^n_{\theta^{-}}[k+1]$ and $o^n_{\theta}[k]$, respectively, $y$ is the index of the maximum element of $o^n_{\theta}[k+1]$, $r^n[k] + \gamma o^n_{y, \theta^{-}}[k+1]$ is the target Q-value.
To stabilize the training targets, the target network is only periodically synchronized with the evaluation network.

The input weights and the recurrent weights of ESNs are randomly initialized according to the constraints specified by the Echo State Property~\cite{luko2012ESN}, and then they remain untrained.
Only the output weights of ESNs are trained so the training is extremely fast.
The main idea of ESNs is to generate a large reservoir that contains the necessary summary of past input sequences for predicting targets.
From Equation (\ref{eqn:hidden_state}), we can observe that the hidden state $h^n[k]$ at any given time slot $k$ is unchanged during the training process if the input weights and recurrent weights are fixed.
In contrast to conventional RNNs that usually initialize the hidden states to zeros and waste some training examples to set them to appropriate values in one training iteration, the benefit of ESNs is that the hidden states do not need to be reinitialized in every training iteration.
Therefore, the training process becomes extremely efficient, which is especially suitable for learning in a high dynamic environment.
Compared to storing $(s[k], a[k], r[k], s[k+1])$ in conventional DRQN framework, we also store hidden states $(h[k], h[k+1])$ because hidden states are unchanged.
In this way, we do not have to waste lots of training time and data to recalculate hidden states in every training iteration.
It largely boosts the training efficiency in the highly dynamic environment since we can avoid using BPTT and only update the output weights of networks.
Furthermore, we can randomly sample from the replay memory to create a training batch, while conventional DRQN methods have to sample continuous sequences to create a training batch.
Thus the training data can be more efficiently used in our DEQN method.
The training data stored in the buffer will be refreshed periodically in order to adapt to the latest environment. 
Therefore, our training method is an online training algorithm that keeps updating the learning agent.
The training algorithm for DEQNs in the DSS problem is detailed in Algorithm~\ref{alg:DEQN_training}.

\section{Performance Evaluation}

\subsection{Experimental Setup}

\begin{figure}[ht]
\centering
	\includegraphics[width=.8\linewidth]{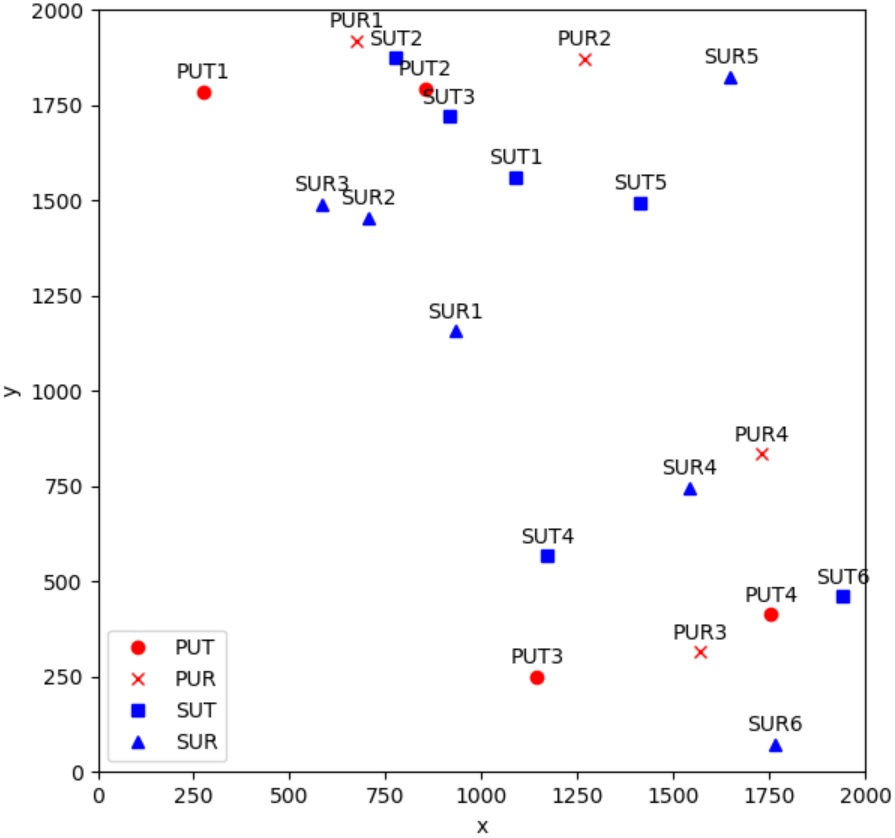}
	\caption{The DSS network geometry. PUT/SUT represent the transmitter of PU/SU. PUR/SUR represent the receiver of PU/SU.}
	\label{fig:geometry}
\end{figure}

We set the number of PUs and SUs to 4 and 6, respectively, and the locations of PUs and SUs are randomly defined in a 2000m$\times$2000m area.
The distance between the transmitter and the receiver of each desired link is randomly chosen from 400m-450m.
Figure~\ref{fig:geometry} shows the geometry of the DSS network, where PUT/SUT represent the transmitters of PU/SU and PUR/SUR represent the receivers of PU/SU.
The channel gains of desired links, interference links, and sensing links are generated by the WINNER II channel model widely used in 3GPP LTE-Advanced and 5G networks~\cite{winner2}.
In this case, there are 4 desired links for PUs, 6 desired links for SUs, 30 interference links between different SUs, 24 interference links between SUTs and PURs, 24 interference links between PUTs and SURs, and 24 sensing links between PUTs and SUTs.
Totally, 112 wireless links are generated in our simulation, which establishes a more complicated scenario than existing DRL-based DSS strategies~\cite{wang2018deep,naparstek2018deep, Chang2019DSA}.
Specifically, \cite{wang2018deep} considers each channel only has two possible states (good or bad) without modeling the true wireless environment; \cite{naparstek2018deep} assumes that the collision between users can be perfectly detected without considering the dynamics of interference links; \cite{Chang2019DSA} assumes that SUs are forbidden to access a channel when a PU is using without considering the actual interference links between PUs and SUs.
For each channel, the bandwidth is set to 5MHz and the variance of the Gaussian noise is set to -157.3dBm. 
The transmit power of PUs and SUs are both set to 500mW.
We set the sensing and transmission period $T$ to 10 time slots and the sensing duration $T_s$ to 2 time slots, where one time slot represents interval of 1ms.
We list all the parameters to generate the wireless environment in Table~\ref{tab:parameters}.

\begin{table}[ht]
\centering
\caption{The values of parameters for generating the wireless environment.}
\label{tab:parameters}  
\begin{tabular}{|c|c|}
\hline
\textbf{Parameter} & \textbf{Value}  \\ \hline
number of PUs $M$ & 4  \\ \hline
number of SUs $N$ & 6  \\ \hline
simulation area & 2000m$\times$2000m \\ \hline
distance between user pair & 400m-450m  \\ \hline
transmit power of PU & 500mW  \\ \hline
transmit power of SU & 500mW  \\ \hline
variance of Gaussian noise & -157.3dBm  \\ \hline
bandwidth of a channel & 5MHz \\ \hline
interval of one time slot & 1ms  \\ \hline
sensing and transmission period $T$ & 10 time slots  \\ \hline
sensing duration $T_s$ & 2 time slots  \\ \hline
\end{tabular}
\end{table}

For the activity pattern of PUs, we let two PUs be in \textit{Active} state every $3T$ (PU1 and PU3) and two PUs be in \textit{Active} state every $4T$ (PU2 and PU4).
Each SU trains its DEQN agent and updates the policy accordingly after collecting 300 samples in the buffer.
The buffer will be refreshed after training so we only use training data from the latest 3 sec.
The total number of training data is 60000, which requires 600 sec to collect all the training data. 
The initial exploration probability $\epsilon$ is set to 0.3, and then it will gradually decrease until $\epsilon$ is 0.
We first train the Q-network with learning rate 0.01, and then the learning rate decreases to 0.001 when $\epsilon$ is less than 0.2.

\begin{figure}[ht]
\centering
\includegraphics[width=1.0\linewidth]{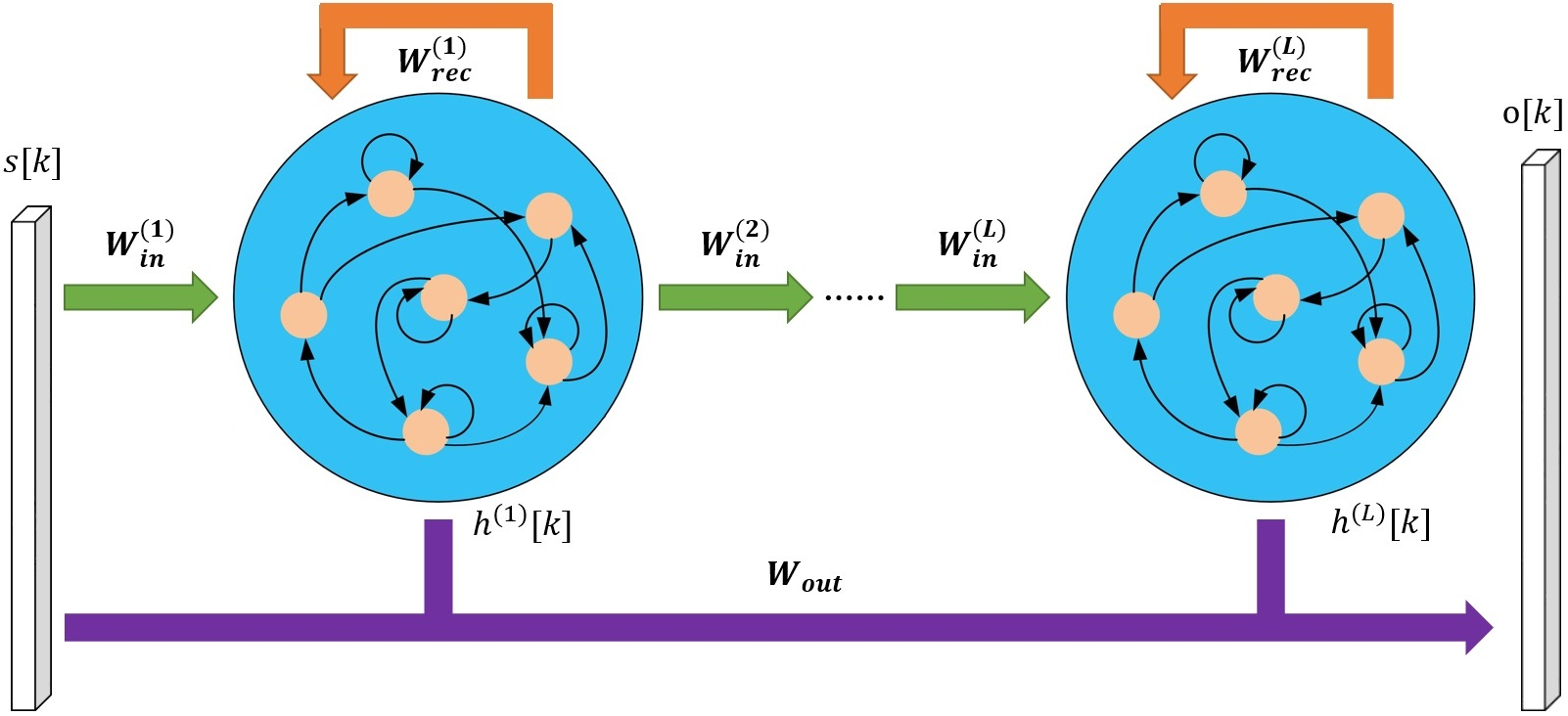}
\caption{The network architecture of DEQN.}
\label{fig:network_architecture}
\end{figure}

\subsection{Network Architecture}
As shown in Figure~\ref{fig:network_architecture}, our DEQN network consists of $L$ reservoirs for extracting the necessary temporal correlation to predict targets.
The number of neurons in each reservoir is set to 32 and the leaky parameter $\beta$ is set to 0.7 in Equation (\ref{eqn:hidden_state}).
During the training process, the input weights $\{ W_{in}^{(1)}, ..., W_{in}^{(L)} \}$ and the output weights $\{ W_{rec}^{(1)}, ..., W_{rec}^{(L)} \}$ are untrained.
To find a good policy, only the output weight $W_{out}$ is trained to read essential temporal information from the input states and the hidden states stored in the experience replay buffer.
Existing research shows that stacking RNNs automatically creates different time scales at different levels, and this stacked architecture has better ability to model long-term dependencies than single layer RNN~\cite{hermans2013DRNN,galli2017deepESN, zhou2020Reservoir}.
We also find that stacking ESNs can indeed improve the performance in our experiment.

\subsection{Results and Discussion}
We evaluate our introduced DEQN method with three performance metrics: 1) The system throughput of PUs. 2) The system throughput of SUs. 3) The required training time.
The throughput represents the number of transmitted bits per second, which is calculated by (spectral-efficiency) $\times$ (bandwidth), and the system throughput represents the sum of users' throughput in the primary system or secondary system.
A good DSS strategy should increase the throughput of SUs as much as possible, while the transmissions of SUs do not harm the throughput of PUs.
Therefore, each SU has to access an available channel by predicting activities of other mobile users. 
We compare with conventional DRQN method that uses Long Short Term Memory (LSTM)~\cite{hochreiter1997LSTM} as the Q-network. 
For a fair comparison, we also set the number of neurons in each LSTM layer to 32.
The training algorithm of DRQNs is BPTT and double Q-learning with the same learning rate as DEQNs. 
Since each SU updates its policy for every 300 samples, we show all of our curves in figures by calculating the moving average of 300 consecutive samples for clarity.

\begin{figure}[ht]
\centering
\includegraphics[width=1.0\linewidth]{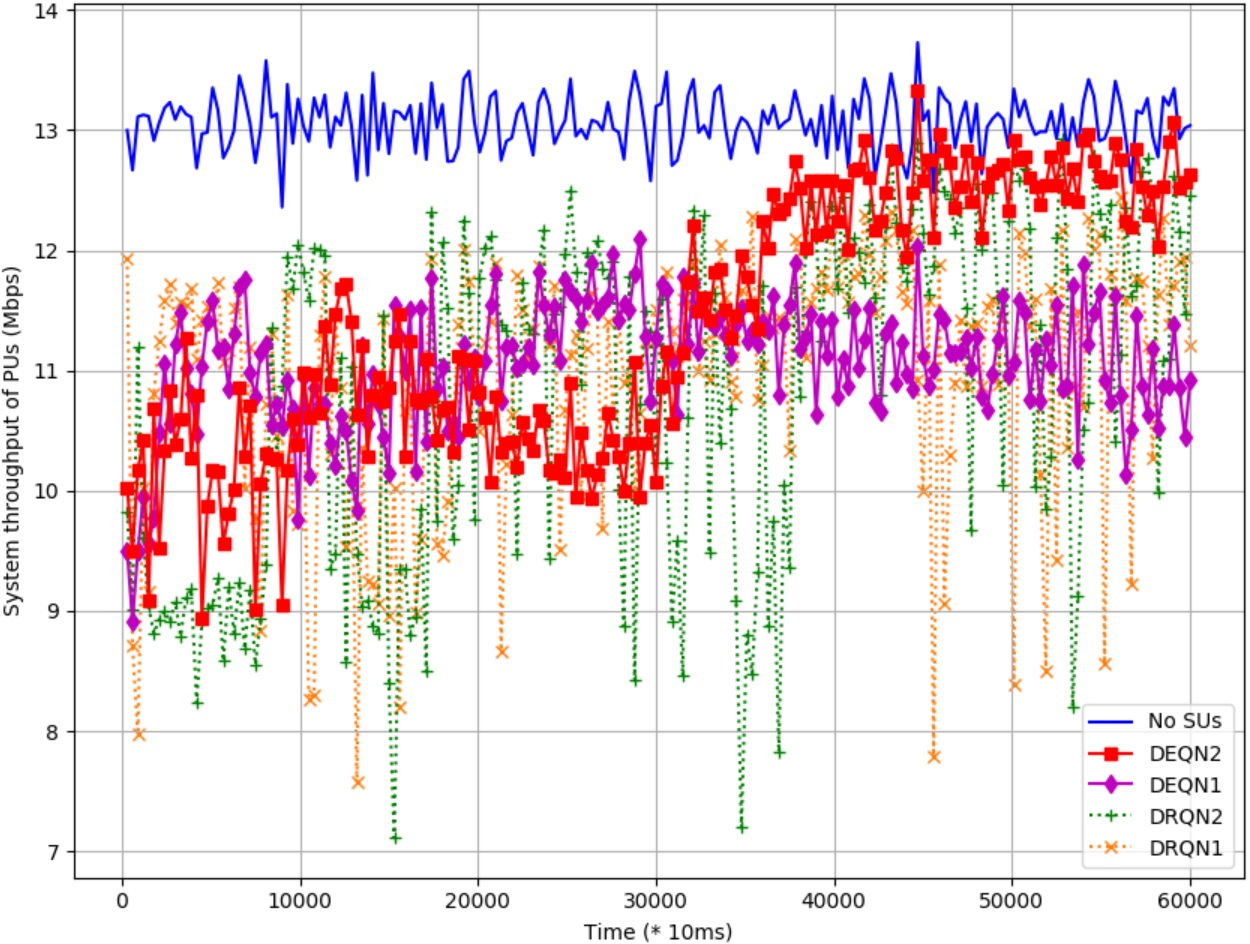}
\caption{The system throughput of PUs.}
\label{fig:PU_throughput}
\end{figure}

\begin{figure}[ht]
\centering
\includegraphics[width=1.0\linewidth]{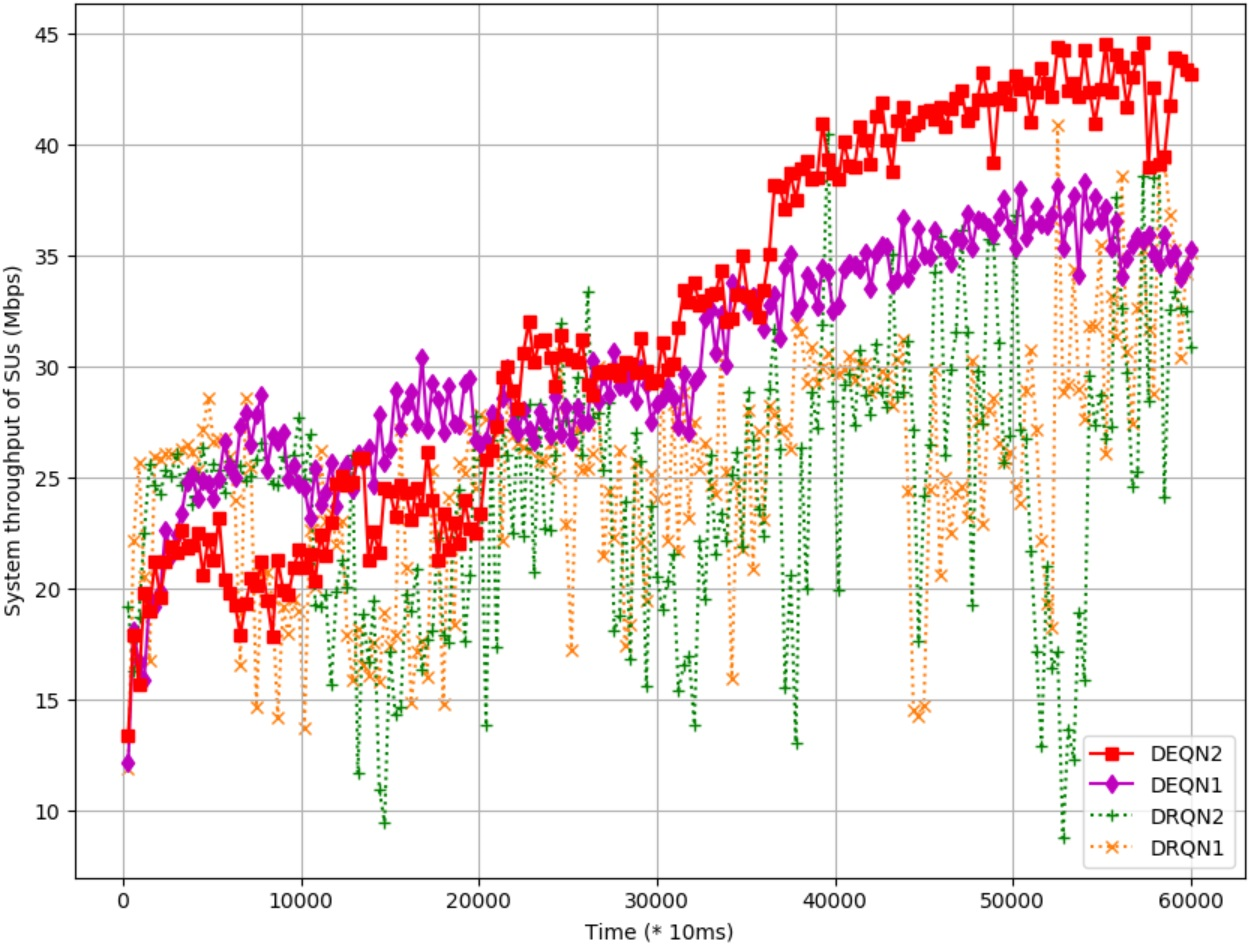}
\caption{The system throughput of SUs.}
\label{fig:SU_throughput}
\end{figure}

\begin{figure}[ht]
\centering
\includegraphics[width=1.0\linewidth]{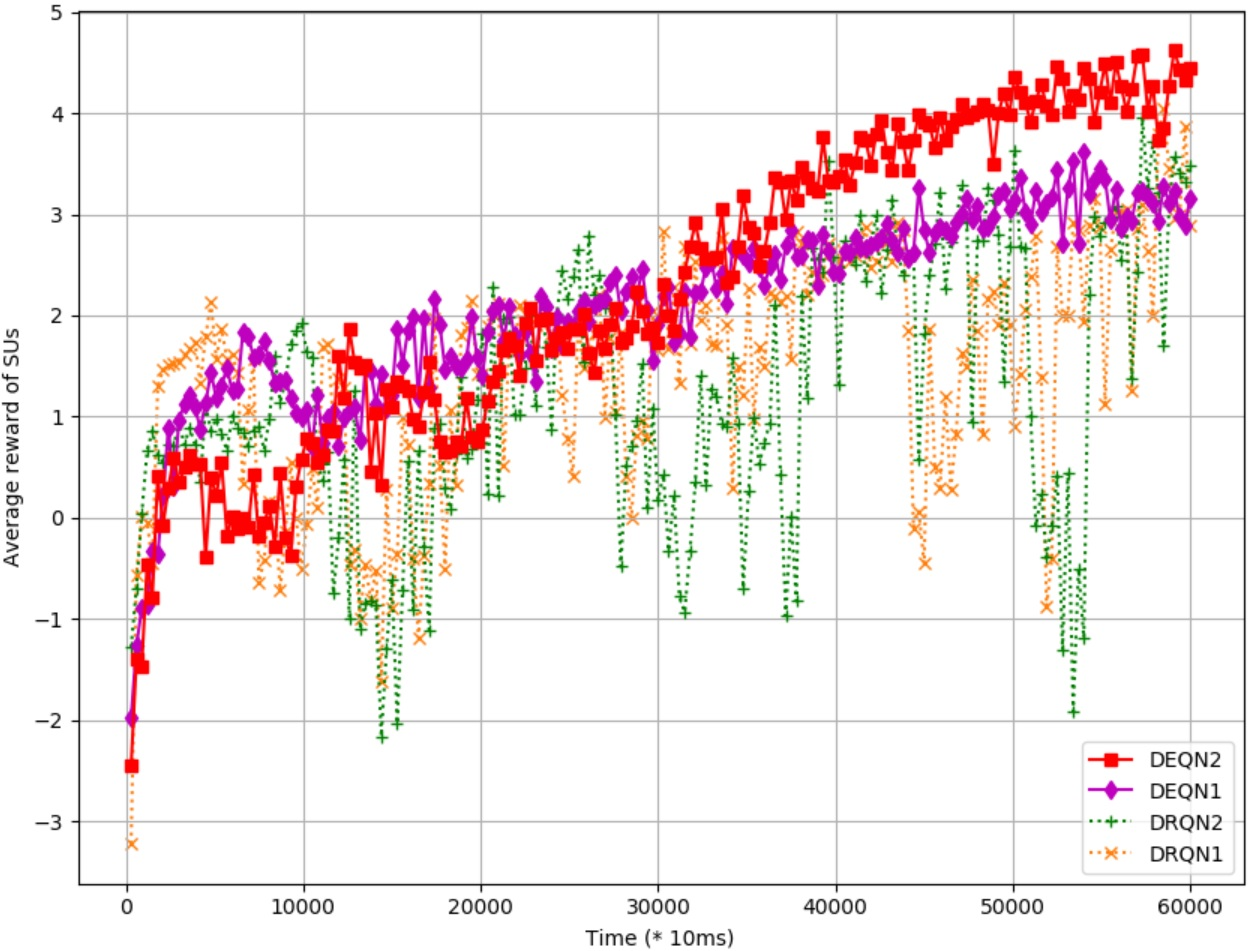}
\caption{The average reward versus time.}
\label{fig:average_reward}
\end{figure}

DEQN1 and DEQN2 are our DEQN method with one and two layers, respectively, and DRQN1 and DRQN2 are the conventional DRQN method with one and two layers, respectively.
The system throughput of PUs is shown in Figure~\ref{fig:PU_throughput} and the system throughput of SUs is shown in Figure~\ref{fig:SU_throughput}.
We observe that DEQNs have more stable performance than DRQNs, which empirically proves that the DEQN method can learn efficiently with limited training data.
Note that one experience replay buffer only contains 300 latest training samples.
After updating the learning agent of each SU using the 300 data in the buffer, DSS strategy of each SU changes so the environment observed by one SU also changes.
Therefore, we have to erase the outdated samples from the buffer and let SUs collect new training data from the environment.
Figure~\ref{fig:average_reward} shows the average reward of SUs versus time.
We observe extremely unstable reward curves of both DRQN1 and DRQN2 so it proves that DRQNs cannot adapt to this dynamic 5G scenario well with few training data.

We observe that DEQN2 has better performance than DEQN1 in both the system throughput of PUs and SUs, which shows that deep structure (stacking ESNs) indeed improves the capability of the DRL agent to learn long-term temporal correlation. 
As for DRQNs, we observe that DRQNs do not have improved performance as we increase the number of layers in the underlying RNN. 
The main reason is that more training data are needed for training a larger network but even DRQN with one layer cannot be trained well.

\begin{figure}[ht]
\centering
\includegraphics[width=0.9\linewidth]{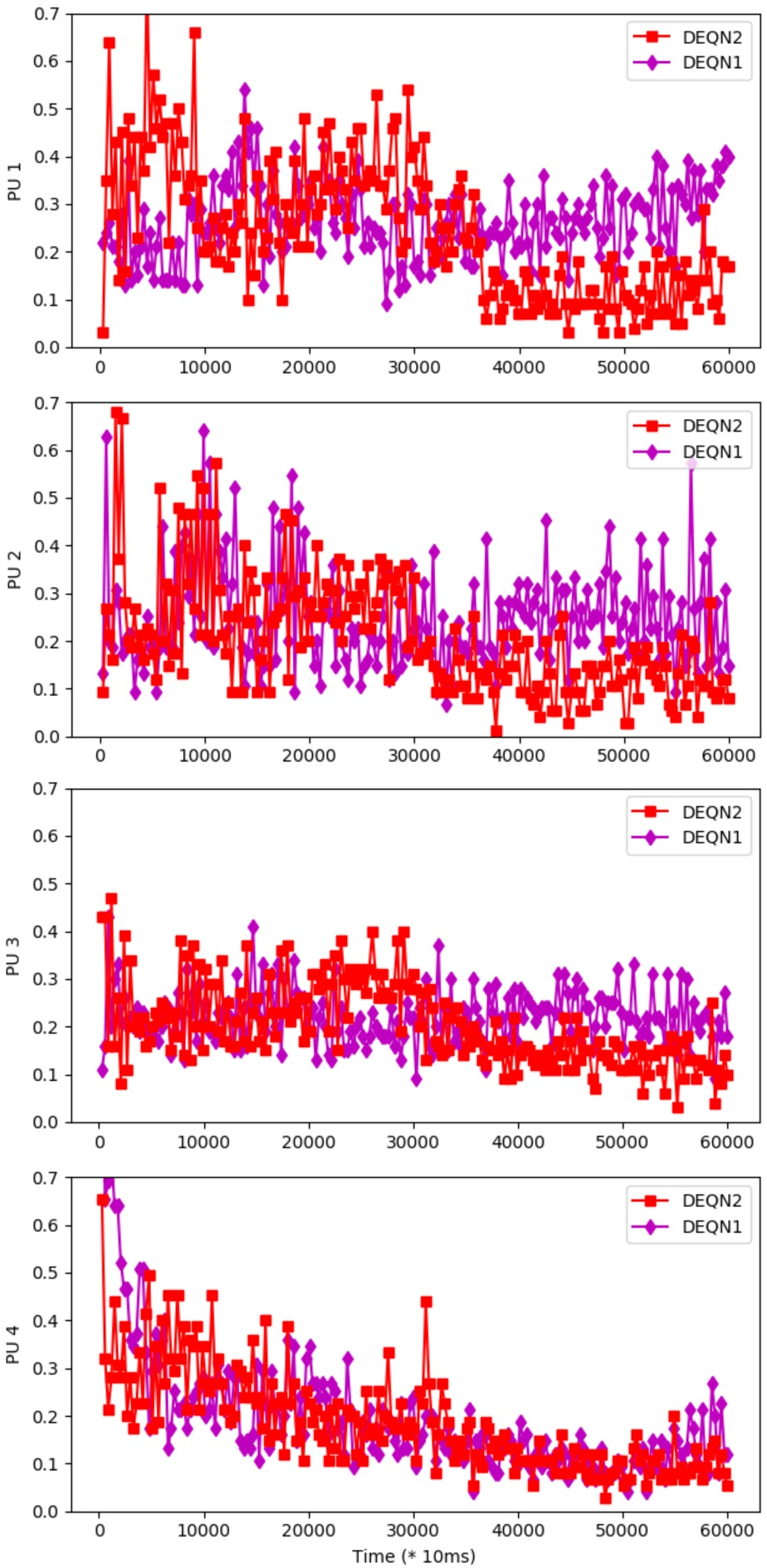}
\caption{The average warning frequency of each PU versus time.}
\label{fig:warning_PU}
\end{figure}

The top priority of designing a DSS network is to prevent harmful interference to the primary system.
To analyze the performance degradation of the primary system after allowing the secondary system to access, we show the system throughput of PUs when there is no SU exist in Figure~\ref{fig:PU_throughput}.
We observe that DEQN2 can achieve almost the same performance of the system throughput of PUs.  
A PU broadcasts a warning signal if its spectral-efficiency is below a threshold.
For each PU, we record the frequency of (the PU sends a warning signal and it is received by some SUs) / (number of the PU's access), which is called as the warning frequency.
Figure~\ref{fig:warning_PU} shows the average warning frequency of each PU versus time.
We observe that the every PU decreases its warning frequency over time, meaning that each SU learns not to access the channel that will cause harmful interference to PUs.

\begin{table}[ht]
\centering
\caption{The comparison of training time of different network architectures.}
\label{tab:training_time}  
\begin{tabular}{|c|c|}
\hline
\textbf{Network} & \textbf{Training time (sec)}  \\ \hline
DEQN1 & 161  \\ \hline
DEQN2 & 178  \\ \hline
DRQN1 & 3776 \\ \hline
DRQN2 & 7618 \\ \hline
\end{tabular}
\end{table}

We compare the training time of different approaches in Table~\ref{tab:training_time} when implemented and executed on the same machine with 2.71 GHz Intel i5 CPU and 12 GB RAM.
The required training time for DRQN1 is 23.4 times the training time for DEQN1, and the required training time for DRQN2 is 42.8 times the training time for DEQN2.
This huge difference shows the training speed advantage of our introduced DEQN method against the conventional DRQN method.
DRQN suffers from high training time because BPTT unfolds the network in time to compute the gradients, but DEQN can be trained very efficiently because the hidden states can be pre-stored for many training iterations.

\section{Conclusion}
In this paper, we introduced the concept of DEQN, a new RNN-based DRL strategy to efficiently capture the temporal correlation of the underlying time-dynamic environment requiring very limited amount of training data.
The DEQN-based DRL strategies largely increase the rate of convergence compared to conventional DRQN-based strategies.
DEQN-based spectrum access strategies are examined in DSS, a key technology in 5G and future 6G networks, showing significant performance improvements over state-of-the-art DRQN-based strategies.
This provides strong evidence for adopting DEQN for real-time and time-dynamic applications.
Our future work will be focused on developing methodologies for the design of neural network architectures tailored to different applications.

%
%

\ifCLASSOPTIONcaptionsoff
  \newpage
\fi

\bibliography{reference.bib}
\bibliographystyle{IEEEtran}

\begin{IEEEbiography}
    [{\includegraphics[width=1in,height=1.25in,clip,keepaspectratio]{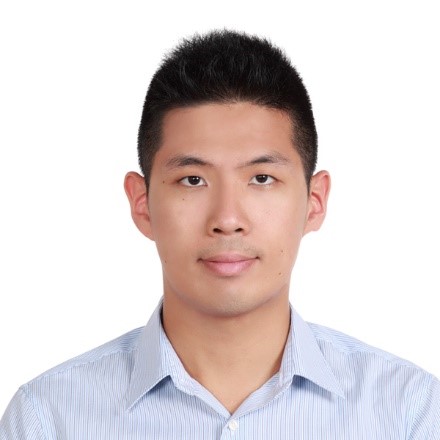}}]{Hao-Hsuan Chang}
received the B.Sc. degree in electrical engineering and the M.S. degree in communication engineering from National Taiwan University, Taipei, Taiwan. He is currently pursuing the Ph.D. degree in electrical and computer engineering at Virginia Polytechnic Institute and State University, Blacksburg, VA, USA. His research interests include dynamic spectrum access, echo state network, and deep reinforcement learning.
\end{IEEEbiography}

\begin{IEEEbiography}
    [{\includegraphics[width=1in,height=1.25in,clip,keepaspectratio]{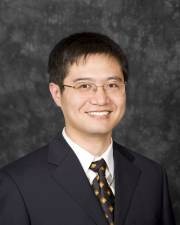}}]{Lingjia Liu}
is an Associate Professor with the Bradley Department of Electrical Engineering and Computer Engineering, Virginia Tech. He is also the Associate Director of Wireless@VT. Prior to joining VT, he was an Associate Professor with the EECS Department, University of Kansas (KU). He spent more than four years working in the Mitsubishi Electric Research Laboratory (MERL) and the Standards and Mobility Innovation Laboratory, Samsung Research America (SRA), where he received the Global Samsung Best Paper Award in 2008 and 2010. He was leading Samsung’s efforts on multiuser MIMO, CoMP, and HetNets in LTE/LTE-advanced standards. His general research interests mainly lie in emerging technologies for beyond 5G cellular networks, including machine learning for wireless networks, massive MIMO, massive MTC communications, and mmWave communications.

He received the Air Force Summer Faculty Fellow, from 2013 to 2017, Miller Scholar at KU, 2014, Miller Professional Development Award for Distinguished Research at KU, 2015, the 2016 IEEE GLOBECOM Best Paper Award, the 2018 IEEE ISQED Best Paper Award, the 2018 IEEE TAOS Best Paper Award, 2018 IEEE TCGCC Best Conference Paper Award, and 2020 WOCC Charles Kao Best Paper Award.
\end{IEEEbiography}

\begin{IEEEbiography}
    [{\includegraphics[width=1in,height=1.25in,clip,keepaspectratio]{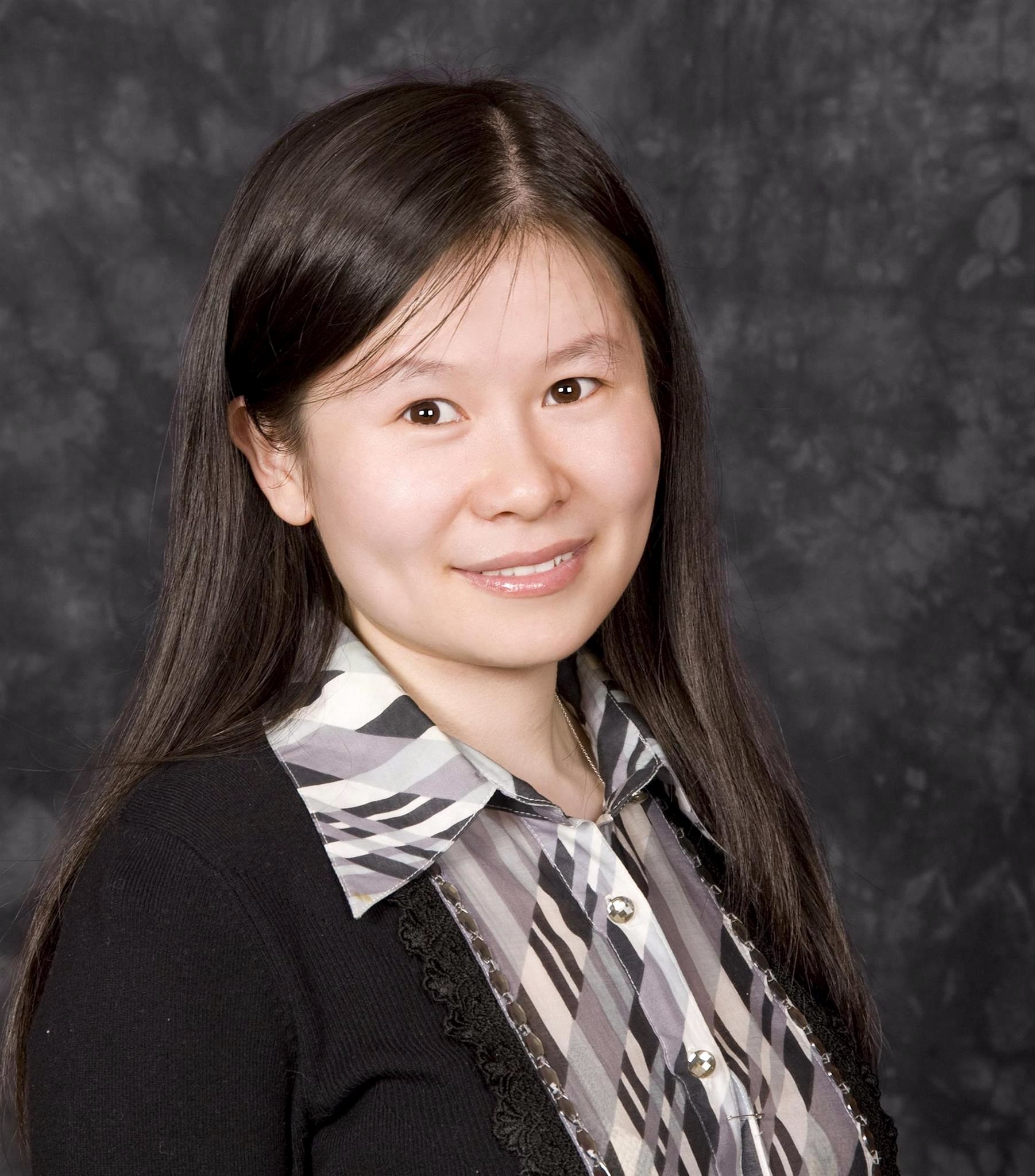}}]{Yang Yi}
(SM’17) is an Associate Professor in the Bradley Department of ECE at Virginia Tech (VT). She received the B.S. and M.S. degrees in electronic engineering at Shanghai Jiao Tong University, and the Ph.D. degree in electrical and computer engineering at Texas A\&M University. Her research interests include very large scale integrated (VLSI) circuits and systems, computer aided design (CAD), and neuromorphic computing. Dr. Yi is currently serving as an associate editor for cyber journal of selected areas in microelectronics and has been serving on the editorial board of international journal of computational \& neural engineering. Dr. Yi is the recipient of 2018 National Science CAREER award, 2016 Miller Professional Development Award for Distinguished Research, 2016 United States Air Force (USAF) Summer Faculty Fellowship, 2015 NSF EPSCoR First Award, and 2015 Miller Scholar.
\end{IEEEbiography}

\end{document}